# High-resolution power equipment recognition based on improved self-attention

Siyi Zhang[1], Cheng Liu[1], Xiang Li[1], Xin Zhai[1], Zhen Wei[1], Sizhe Li[2], Xun Ma[2]
Jinan Power Supply Company, Jinan, Shandong, China, 250000
Nanjing University Of Finance & Economics, Nanjing, Jiangsu, China, 210000

Abstract: The current trend of automating inspections at substations has sparked a surge in interest in the field of transformer image recognition. However, due to restrictions in the number of parameters in existing models, high-resolution images can't be directly applied, leaving significant room for enhancing recognition accuracy. Addressing this challenge, the paper introduces a novel improvement on deep self-attention networks tailored for this issue. The proposed model comprises four key components: a foundational network, a region proposal network, a module for extracting and segmenting target areas, and a final prediction network. The innovative approach of this paper differentiates itself by decoupling the processes of part localization and recognition, initially using low-resolution images for localization followed by high-resolution images for recognition. Moreover, the deep self-attention network's prediction mechanism uniquely incorporates the semantic context of images, resulting in substantially improved recognition performance. Comparative experiments validate that this method outperforms the two other prevalent target recognition models, offering a groundbreaking perspective for automating electrical equipment inspections.

## 1 Introduction

In recent years, as the scale of the power grid continues to expand, the contradiction between the number of equipment and inspection staffing has become increasingly prominent. At present, the mainstream inspection method in high-voltage substations is manual work, which is not only time-consuming and laborious, but also the inspection quality cannot be guaranteed in bad weather. Therefore, more and more research has begun to focus on automated inspection technology, compared with manual inspection, it can integrate a variety of inspection methods, high efficiency and stability. With the development of hardware technologies such as robots and drones and software technologies such as artificial intelligence and big data analysis, substation automated inspection technology that automatically collects images of equipment, analyzes and reports safety hazards through inspection robots has become possible, which reduces the workload of station operation and maintenance while improving the level of operation and maintenance [1,2]. However, in this process, the key problem of component identification still needs to be studied: at present, limited by the number of model parameters, it is impossible to directly use high-resolution inspection images, and there is a large room for improvement in the recognition accuracy.

At present, domestic and international research on image recognition of substation equipment mainly uses the existing model framework in the computer field, and does not consider the problem of high resolution. Literature [3] for the detection of foreign objects in the power grid, proposed a sample expansion method, although the increase in the amount of data makes the model effect to improve, which directly uses the existing Faster RCNN model, and the input image is a reduced image. Literature [4] proposed a new feature for the substation image recognition problem by fusing image sift features with sparse representation and using a classification model to determine the equipment category. The limitation of this method is that it cannot deal with complex images containing multiple devices, and it is extremely sensitive to external conditions such as shooting angle and illumination. Literature [5] proposes a grayscale-based coding method for the transformer localization and identification problem, using which image matching can be achieved. The template matching method has poor robustness and is completely unable to adapt itself to different resolution images. In summary, existing studies rarely focus on the resolution problem in transformer equipment recognition, and most of them directly use image recognition algorithms in the computer field, which makes the existing methods perform poorly in small-size part recognition, similar type part recognition, and other scenarios that require high resolution.

This paper aims to solve the key problem in transformer inspection: high-resolution image recognition, and proposes an improved deep self-attention network model for this scenario, which uses the scaled-down image to determine the location of the part, and then extracts the features of the corresponding region in the high-resolution original image, and utilizes the deep self-attention network for recognition. The method retains the features of the part image at high resolution while ensuring the operation speed, thus improving the recognition accuracy. Experiments show that this method is significantly better than the remaining two commonly used target recognition models (Faster RCNN network and SSD network), which is of guiding significance for the realization of inspection automation in this scenario.

## 2 Network Architecture

### 2.1 Network Architecture

Figure 1 gives the structure of the improved deep self-attention network, which contains four main parts: the base network, the region candidate network, the target region extraction and segmentation module, and the prediction network. The base network adopts the ResNet network [6], which aims to extract features from the equally scaled down image, and the advantages of using the scaled down image are to reduce the size of the base network, increase the speed of the model operation, and reduce the dependence of the model on the computational resources. The ResNet network will extract the multi-level features, which will be combined in a pyramid structure [7], and be used as inputs for the region candidate network. . The purpose of the region candidate network [8] is to determine the location of the target object based on the image features, this step only distinguishes between the foreground and the back view, not the type of parts, and since all parts are considered as one class, i.e., the foreground, together with the downsampling of the back view, the imbalance in the number of foreground and back view samples has been improved and the accuracy of the model is greatly improved. Up to this point, the part locations are determined, and the next step is to determine the part types.

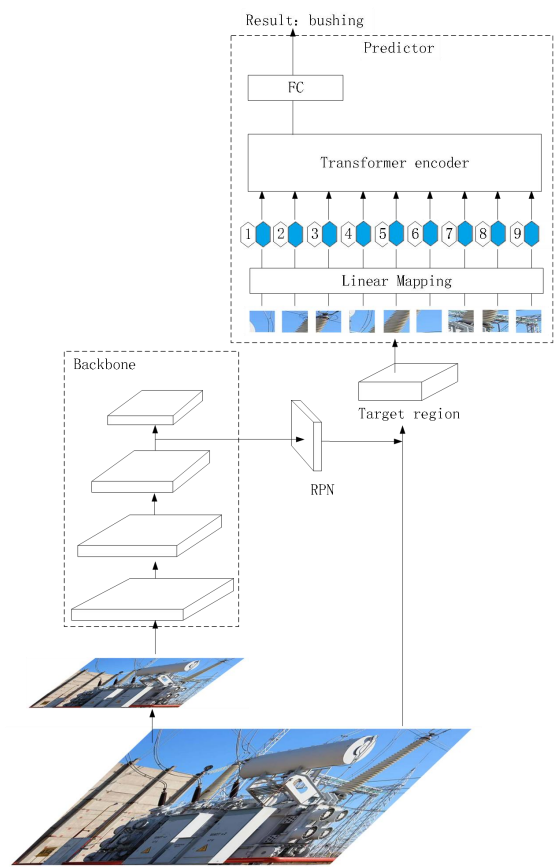

Fig.1 Modified deep self-attention network structure

Different from the traditional target recognition neural network model, high resolution images are used here for part category recognition. Traditional target recognition networks such as Faster RCNN [8] directly use the features of the base network for category recognition, which is not conducive to the recognition of small size parts. There are the following contradictions: inputting high-resolution images into the base network can recognize small-sized parts, but the model operation occupies a lot of computational resources, is slow, and is difficult to adjust the parameters; inputting reduced images into the base network, the model operation occupies less resources, the speed is greatly improved, and it is easy to adjust the parameters; however, small-sized parts occupy only a few pixels in the reduced image, which makes it difficult to recognize them. Therefore, the method proposed in this paper firstly utilizes the reduced image to determine the position of the part, and then inputs the high-resolution image into the network to ensure the recognition effect. Here is the corresponding relationship between position $(x', y')$ on the reduced image and position $(x, y)$ on the high-resolution original image:

$$x = \frac{w}{w'}x', \quad y = \frac{h}{h'}y' \quad (1)$$

where $w, h$ is the width and height of the high-resolution original image and $w', h'$ is the width and height of the reduced image.

The output result of the above region candidate network is a rectangular box containing the components, where are the coordinates of the center of the rectangular box and are the width and height of the rectangular box. The output result of the candidate network in the above area is a rectangular frame $(x'_i, y'_i, w'_i, h'_i)$ containing components, where $(x'_i, y'_i)$ is the center coordinate of the rectangular frame and $w'_i, h'_i$ is the width and height of the rectangular frame. Using formula (1), the position $(x_i, y_i, w_i, h_i)$ of the corresponding component on the high-resolution original image can be obtained. It should be noted that the candidate box for eliminating repeated prediction [9] by maximum suppression is used here. These information will be used as the input of the region extraction and segmentation module, and the corresponding output will be the segmented region. The specific segmentation method is as follows: firstly, the pixel matrix corresponding to the rectangular frame $(x_i, y_i, w_i, h_i)$ is scaled to a fixed scale of $l \times l$, and then the matrix is divided into $N$ equal parts $D_1, \cdots, D_N$. This segment-ation method is adaptive, for large-size parts such as transformers, which can be recognized with a low resolution, and which occupy a large number of pixel points on the original image, the scaling ratio is larger when transforming to a fixed scale; while for small-size parts such as insulators, which require a high resolution to be recognized, and which occupy only a small number of pixel points on the original image, the scaling ratio is smaller when transforming to a fixed scale, and most of the high-resolution is retained.

The images of the above $N$ regions will be fed into the prediction network, which will give the category and confidence level of the parts, here the prediction network is based on the deep self-attention network [10], which is divided into four parts: region linear mapping, fixed position coding, encoder, and fully connected layer. Compared to other structures, the deep self-attention network takes into account the inter-region relationships and the overall semantics of the image and has a higher recognition accuracy.

2.2 Prediction Network

This chapter will introduce the prediction network in detail, which is partly based on the deep self-attention network and contains the structures of region linear mapping, fixed position encoding, encoder, and fully connected layer. First is the region linear mapping layer, the purpose of this module is to map the original image to the feature space, the specific details are as follows: The location of an object in the original high-resolution image is given by a rectangular frame $(x_i, y_i, w_i, h_i)$. The pixel matrix is scaled to a fixed size and divided into $N$ regions $D^i_j \in R^{P \times P \times C}$, where $i$ is the part number, $1 \leq j \leq N$ is the region number, $P = l/\sqrt{N}$ is the region size, $C$ is the number of channels, and for RGB images, $C = 3$. The number of neurons in the linear mapping layer is $d$, and the output of each neuron is the weighted sum of each channel in each position in the region, where the weight matrix is $W \in R^{CP^2 \times d}$, so the linear mapping can be expressed as $D^i_j W \in R^{1 \times d}$. Here, the component category code $x_{class}$ and the region position code $E_{pos}$ [10] are also embedded in the feature vector, that is, the feature vector is embedded in the feature vector.

The linear mapping layer contains the number of neurons as, the output of each neuron is the weighted sum of each channel at each location in the region, where the weight matrix, so the linear mapping can be expressed as. Here the part type coding and region location coding [10] are also embedded in the feature vector, i.e.

$$z^i_o = [x_{class}; D^i_1 W; \cdots; D^i_N W] + E_{pos} \quad (2)$$

Where $x_{class} \in R^{1 \times d}$ and $E_{pos} \in R^{(N+1) \times d}$ are learnable parameters, and $z^i_o \in R^{(N+1) \times d}$ is the feature vector of the $i$ component input encoder.

The encoder, on the other hand, uses the standard encoder for deep self-attention networks [11], and assuming that the number of its layers is $L$, and each layer contains two modules: the multi-branch self-attention module MSA, and the fully-connected layer MLP, the input-output

relationship of the encoder can be expressed as follows:

$$z_\ell^{i\prime} = \text{MSA}(\text{LN}(z_{\ell-1}^i)) + z_{\ell-1}^i \quad (3)$$

$$z_\ell^i = \text{MLP}(\text{LN}(z_\ell^{i\prime})) + z_\ell^{i\prime} \quad (4)$$

$$y^i = \text{LN}(z_L^i[0]) \quad (5)$$

where LN denotes the normalization layer, where the normalization direction is the channel direction. $z_\ell^i$ represents the output of the MSA module on the $\ell(1 \leq \ell \leq L)$ floor corresponding to the $i$ component, and $z_\ell^{i\prime}$ represents the output of the MLP module on the $\ell$ floor corresponding to the $i$ component, where both MSA module and MLP module adopt residual structure and are connected in series alternately. Fig. 2 shows the schematic structure of the encoder. $y^i$ indicates the output of the encoder corresponding to the $i$ component, and $z_L^i[0]$ indicates the magnitude of the MLP output of the layer $L$ at position 0, which corresponds to the component type code at the beginning, namely $x_{class}^i = z_o^i[0]$.

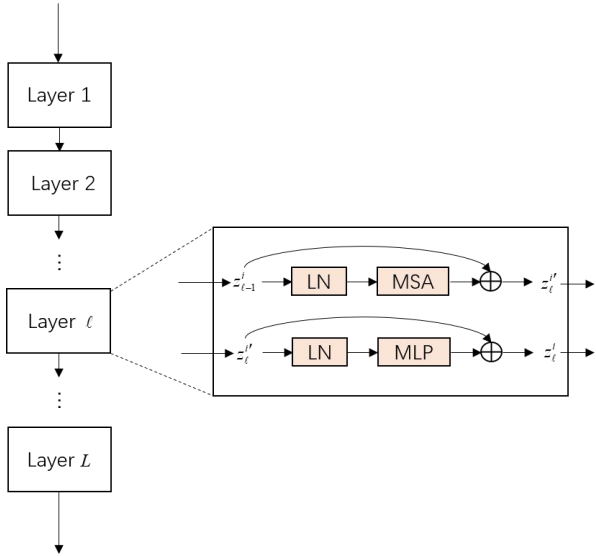

Fig.2 Encoder structure diagram

At this time, the result $y^i \in R^{1 \times D}$ output by the encoder is the encoded feature vector, and the probability $p_c^i$ that the component belongs to each category can be obtained by using the last fully connected layer $W_c$.

$$p_c^i = \text{softmax}(f(y^i W_c)) \quad (6)$$

where $W_c \in R^{D \times N_c}$ is the weight of the fully connected layer, $N_c$ is the number of part categories, $f$ is the sigmoid activation function, after which the result is converted to probability form using the softmax function.

2.3 Error function

High-resolution self-attention network belongs to the two-stage target recognition model, i.e., to determine the location of foreground objects first, and then to determine the foreground object categories, so it is necessary to design two error functions, which are used in the region candidate network and the prediction network, respectively.

First is the region candidate network, whose error function is as follows:

$$L_{RPN} = \frac{1}{N_{cls}} \sum_k L_{cls}(t_k^c, t_k^{c*}) + \lambda \frac{1}{N_{reg}} \sum_k t_k^{c*} L_{reg}(t_k^p, t_k^{p*}) \quad (7)$$

where $N_{cls}$ is the corresponding category probability of the candidate box, $N_{reg}$ is the corresponding number of position parameters of the candidate box, assuming that the number of candidate boxes is $K$, that is $N_{cls} = 2K$, the probability that each candidate box belongs to the front and back view respectively, that is $N_{reg} = 4K$, each candidate box is represented by four position parameters; $\lambda$ error trade-off parameters, that is, the trade-off between the category error and the position error; $t_k^c$ and $t_k^p$ are the predicted category and the position parameter of the $k$ candidate box, respectively; $t_k^{c*}$ and $t_k^{p*}$ are the true category and the position parameter of the $k$ candidate box, respectively; $L_{cls}$ is the categorization error, i.e., the cross-entropy between two sets of probabilities [14], here for a binary classification problem; and $L_{reg}$ is the location error, i.e., the smooth L1 error [15].

Next is the prediction network with the following error function:

$$L_{TR} = \frac{1}{N_{eb}} \sum_i L_{cls}(p_c^i, p_c^{i*}) \quad (8)$$

where $N_{eb}$ is the number of component bounding boxes, $p_c^i$ is the prediction probability that component $i$ belongs to each category, and $p_c^{i*}$ is the labeling information of component $i$. and $L_{cls}$ is the classification error, i.e., the cross-entropy between the two sets of probabilities, and here is a multi-classification problem.

## 3 Experimental Validation

In this section, the improved deep self-attention network is compared with two other commonly used target recognition methods for the transformer recognition task. The image sources in the dataset include on-site shooting and network collection, of which 120 images are shot on-site and 80 images are collected on the network, and the size of the whole dataset is 200. For the on-site shooting images, the original resolution of the images is 1920×1080; for the images collected on the network, the original resolution of the images is not fixed, and the resolution is unified to be increased to the same size by using the AI image upscaler. to the same size. For the basic network, it uses the equal scale reduced image, the resolution is 640×360. all the images are manually labeled, and the ratio of the training set to the test set is 4:1.

3.1 Improved Deep Self-Attention Network

Due to the small dataset used in this paper, both the base network and the prediction network use a pre-trained model based on the MS COCO dataset, on which the dataset of this paper is utilized for fine-tuning. Among them, different parts correspond to different regions on the original image, and there are differences in the features of the pixels in each region, the prediction network firstly chunks the above regions, and inputs the deep self-attention network encoder by using the linear mapping of the regions, and the encoded features have obvious distributional differences in their feature space, and at this time, the probability that the region belongs to different parts can be calculated by using the fully connected layer. Figure 3 shows the experimental results of the method proposed in this paper, and the bounding box of each component is given in the figure. It can be seen that the method proposed in this paper can accurately recognize various types of parts, and because the method uses high-resolution images, the phenomenon of missed detection hardly occurs.

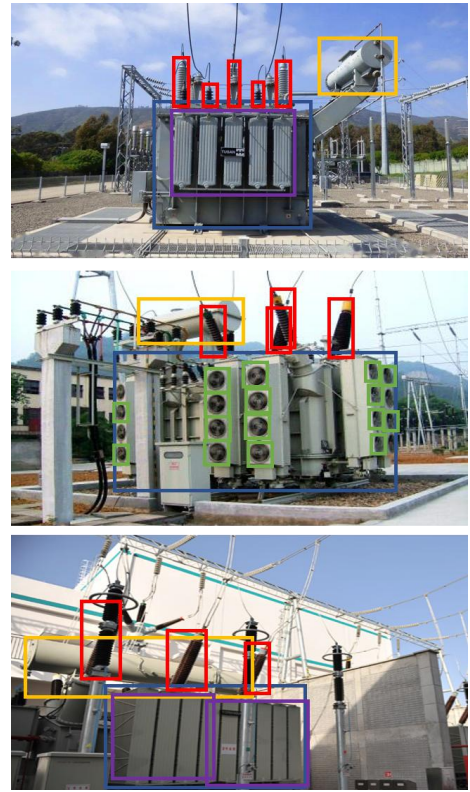

Fig.3 Experimental results of modified deep self-attention network

In the task of target recognition, the average accuracy [16] is the most commonly used evaluation index, and the judgment criteria contain two parts: category and location. The category judgment only needs to consider the confidence level of the corresponding category. The most commonly used metric for location judgment is IoU, i.e., the ratio of the intersection and concatenation of two rectangular boxes [17]. The higher the IoU, the closer the two rectangular boxes are in terms of location and size. Different accuracy and recall can be obtained by varying the confidence threshold. The average accuracy is defined as the mean of the accuracy at different recall rates.

In order to explore the optimal network structure, the accuracy under different network structures was tested and the test results are shown in Table 1. The whole test is divided into four stages:

1) Number of ResNet network layers. Here 50 and 101 layers of network are used respectively, which are denoted by R50 and R101 in the table. It can be seen that the model recognition accuracy decreases instead when the number of ResNet network layers increases. This is due to the small amount of experimental data, which is not enough to support deeper layers of the network.

2) Feature pyramid network. The effect of

whether to use a pyramid network on the accuracy is tested here, which is denoted by FPN in the table, and it can be seen that the use of a pyramid network improves the model recognition accuracy significantly.

Tab.1 The influence of network structure on accuracy

| Model | mAP | AP50 | AP75 |
|---|---|---|---|
| R50 | 0.783 | 0.843 | 0.796 |
| R101 | 0.760 | 0.817 | 0.774 |
| R50+FPN | 0.783 | 0.843 | 0.796 |
| R50 without FPN | 0.721 | 0.839 | 0.735 |
| RPN+NMS | 0.783 | 0.843 | 0.796 |
| RPN without NMS | 0.719 | 0.790 | 0.734 |
| T16 | 0.783 | 0.843 | 0.796 |
| T32 | 0.715 | 0.774 | 0.725 |

3) Non-great value suppression. The effect of whether or not to use great magnitude suppression on the accuracy is tested here and is represented by NMS in the table, and it can be seen that using great magnitude suppression improves the model accuracy substantially.

4) Number of layers of deep self-attentive network encoder. Here, 16-layer and 32-layer encoders are tested, which are denoted by T16 and T32 in the table, and it can be seen that the performance of the model decreases instead of increasing the number of layers, indicating that the size of the dataset used in this experiment is not large enough to support deeper layers of encoders.

At this point, the optimal network structure has been determined: the base network is a 50-layer ResNet network with a feature pyramid network, the region candidate network is followed by the addition of non-maximal suppression, and the number of layers of the deep self-attentive network encoder network is set to 16.

3.2 Comparison with other models

As typical target recognition networks, Faster RCNN and SSD [18] are often used as comparative benchmarks to measure the performance of the algorithms. The above two models represent two kinds of target recognition ideas: 1) generating object location candidate frames first, and then performing classification and position regression on the candidate frames; and 2) sampling at different locations of the original image and directly utilizing the corresponding features for classification and position regression.

The commonality between the Faster RCNN network and the model proposed in this paper is that both belong to a two-stage network, i.e., the object candidate frames are determined first, and then the object categories are determined. The differences are twofold: 1) because the base network and the regional feature extraction layer in Faster RCNN use the same feature map, the number of network parameters corresponding to the input high-resolution image is too large, and the network often uses the reduced image as input, resulting in a low recognition rate of small objects; 2) the prediction branch in Faster RCNN only uses the fully connected layer, without considering the semantic

The target recognition accuracies of different models are given in Table 2, where HRTR denotes the method proposed in this paper, and the recognition accuracies of different kinds of parts are given here in detail. It can be seen that the recognition rate of the body is generally high, and the recognition rate of the casing and fan is generally low, which is due to the fact that the size of the body is much larger than the rest of the parts, which makes it less difficult to recognize; whereas the casing and fan are smaller in size and easy to be blocked, which makes it more difficult to recognize. In addition, the proposed method in this paper increases the recognition rate of all types of parts compared to Faster RCNN and SSD networks, and the part categories with the largest improvement are casing and fan. This is due to the fact that in high-resolution images, the input features are significantly increased and the model can learn more detailed information.

Tab.2 Target recognition accuracy of different models

| Model / Type | Mean Accuracy Rate (mAP) | | |
|---|---|---|---|
| | *HRTR* | *Faster RCNN* | *SSD* |
| Ontology | 0.847 | 0.821 | 0.792 |
| Sleeve | 0.740 | 0.649 | 0.610 |
| Oil Pillow | 0.794 | 0.741 | 0.731 |
| Heat Sink | 0.771 | 0.704 | 0.687 |
| Fan | 0.761 | 0.680 | 0.664 |
| **Average** | 0.783 | 0.719 | 0.697 |

Combining the experimental results of the three models and the comparison of target recognition accuracy, it is easy to find that the method proposed in this paper adopts high-resolution images as the input of the prediction network, which can effectively improve the performance of transformer part recognition, especially for some kinds of parts with small size and high demand for recognition details.

## 4 Conclusion

In this paper, an improved deep self-attention network is proposed to address the problem of not being able to use high-resolution images in transformer part image recognition scenarios. The network contains four parts: the base network, the region candidate network, the target region extraction and segmentation module, and the prediction network. The base network is used to extract features from the reduced image, and the region candidate network predicts the part location based on the features, after which the target region extraction and segmentation module obtains the corresponding location information of the high-resolution original image and segments it, and finally gives it to the prediction network to output the part category. Using the above model structure, the method proposed in this paper is significantly better than the remaining two commonly used target recognition methods: the Faster RCNN and the SSD network in the experiments, which provides a new way of thinking to realize the automation of power equipment inspection.